\documentclass[letterpaper]{article}

\newif\ifcomments
\commentsfalse

\usepackage{times}
\usepackage{graphicx} 
\usepackage{subfigure}

\usepackage{natbib}

\usepackage{algorithm}
\usepackage{algorithmic}

\usepackage{hyperref}

\usepackage[accepted]{icml2017}

\usepackage{amsmath,amsthm,verbatim,amssymb,amsfonts,amscd,mathrsfs}
\usepackage[inline]{enumitem}
\usepackage{color}
\usepackage{wrapfig}
\usepackage{booktabs}

\renewcommand{\P}{\mathop{\mathbb{P}}}
\newcommand{\E}{\mathop{\mathbb{E}}}

\newcommand{\scrL}{\mathcal{L}}

\newcommand{\scrX}{\mathcal{X}}
\newcommand{\scrY}{\mathcal{Y}}

\newcommand{\xb}{\bar{X}}

\newcommand{\ind}{\mathbf{1}}

\newcommand{\sep}{\;|\;}

\usepackage{mathtools}

\theoremstyle{plain}

\newtheorem{claim}{Claim}

\theoremstyle{remark}

\renewcommand{\xb}{\mathbf{x}}
\newcommand{\zb}{\mathbf{z}}

\newcommand{\yh}{\hat{Y}}
\newcommand{\ph}{\hat{P}}
\DeclareMathOperator*{\argmax}{argmax}
\DeclareMathOperator{\acc}{acc}
\DeclareMathOperator{\conf}{conf}

\ifcomments\newcommand{\comments}[1]{#1}\else\newcommand{\comments}[1]{}\fi
\definecolor{clrgp}{rgb}{.9,0,.9}

\icmltitlerunning{}
\newcommand{\titl}{On Calibration of Modern Neural Networks}
\newcommand{\titlshort}{On Calibration of Modern Neural Networks}
\newcommand{\authorinfo}{
  \icmlsetsymbol{equal}{*}
  \begin{icmlauthorlist}
    \icmlauthor{Chuan Guo}{equal,cornell}
    \icmlauthor{Geoff Pleiss}{equal,cornell}
    \icmlauthor{Yu Sun}{equal,cornell}
    \icmlauthor{Kilian Q. Weinberger}{cornell}
  \end{icmlauthorlist}

  \icmlaffiliation{cornell}{Cornell University}
  \icmlcorrespondingauthor{Chuan Guo}{cg563@cornell.edu}
  \icmlcorrespondingauthor{Geoff Pleiss}{geoff@cs.cornell.edu}
  \icmlcorrespondingauthor{Yu Sun}{ys646@cornell.edu}
  \icmlkeywords{calibration, confidence, deep learning, neural networks}
}

\allowdisplaybreaks

\begin{document}

\twocolumn[
  \icmltitle{\titl}
  \authorinfo
  \vskip 0.3in
]

\printAffiliationsAndNotice{\textsuperscript{*}Equal contribution, alphabetical order.}

\begin{abstract}
Confidence calibration -- the problem of predicting probability estimates representative of the true correctness likelihood -- is important for classification models in many applications. We discover that modern neural networks, unlike those from a decade ago, are poorly calibrated. Through extensive experiments, we observe that depth, width, weight decay, and Batch Normalization are important factors influencing calibration. We evaluate the performance of various post-processing calibration methods on state-of-the-art architectures with image and document classification datasets. Our analysis and experiments not only
offer insights into neural network learning, but also provide a simple and straightforward recipe for practical settings: on most datasets, \emph{temperature scaling} -- a single-parameter variant of Platt Scaling -- is surprisingly effective at calibrating predictions.
\end{abstract}

\section{Introduction}
\label{introduction}
\setlength{\abovedisplayskip}{4pt}
\setlength{\belowdisplayskip}{4pt}
\setlength{\textfloatsep}{4pt}

Recent advances in deep learning have dramatically improved neural network accuracy \citep{simonyan2014very, srivastava2015highway, he2015deep, huang2016deep, huang2016densely}. As a result, neural networks are now entrusted with making complex decisions in applications, such as object detection \cite{girshick2015fast}, speech recognition \cite{hannun2014deep}, and medical diagnosis \cite{caruana2015intelligible}.
In these settings, neural networks are an essential component of larger decision making pipelines.

In real-world decision making systems, classification networks must not only be accurate, but also should indicate when they are likely to be incorrect.
As an example, consider a self-driving car that uses a neural network to detect pedestrians and other obstructions \cite{bojarski2016end}. If the detection network is not able to confidently predict the presence or absence of immediate obstructions, the car should rely more on the output of other sensors for braking.
Alternatively, in automated health care, control should be passed on to human doctors when the confidence of a disease diagnosis network is low  \cite{jiang2012calibrating}.
Specifically, a network should provide a \emph{calibrated confidence} measure in addition to its prediction.
In other words, the probability associated with the predicted class label should reflect its ground truth correctness likelihood.

\begin{figure}[t!]
	\centering
	\includegraphics[width=\columnwidth]{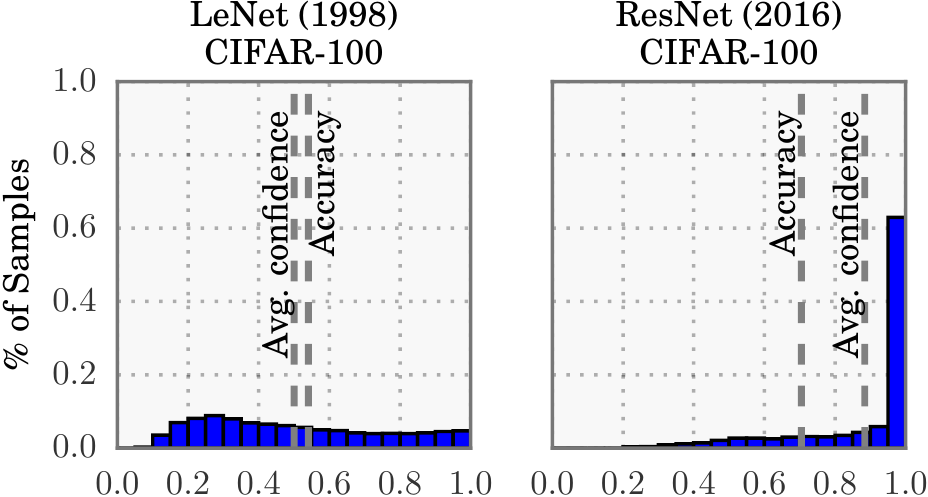}
	\includegraphics[width=\columnwidth]{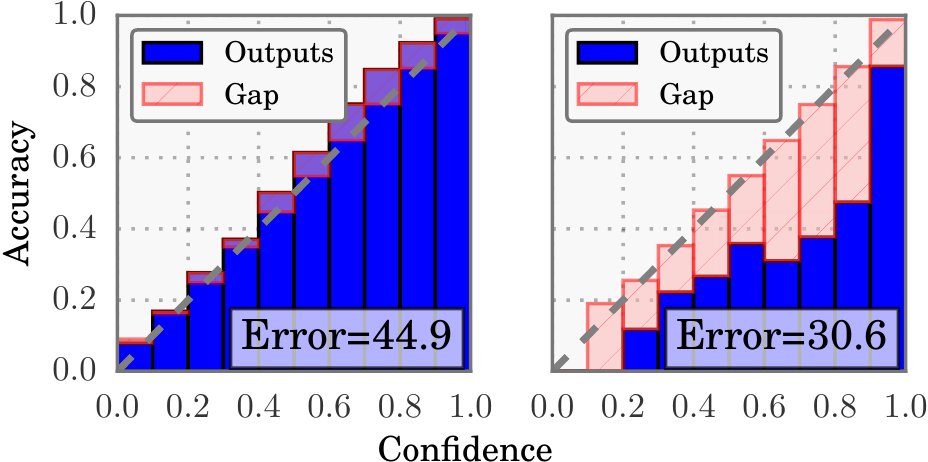}
	\caption{Confidence histograms (top) and reliability diagrams (bottom) for a 5-layer LeNet (left) and a 110-layer ResNet (right) on CIFAR-100. Refer to the text below for detailed illustration.}
	\label{figure.complenet}
	\vspace{8pt}
\end{figure}

Calibrated confidence estimates are also important for model interpretability. Humans have a natural cognitive intuition for probabilities \cite{cosmides1996humans}. Good confidence estimates provide a valuable extra bit of information to establish trustworthiness with the user -- especially for neural networks, whose classification decisions are often difficult to interpret.
Further, good probability estimates can be used to incorporate neural networks into other probabilistic models. For example, one can improve performance by combining network outputs with a language model in speech recognition \cite{hannun2014deep,xiong2016achieving}, or with camera information for object detection \citep{kendall2015modelling}.

In 2005, \citet{niculescu2005predicting} showed that neural networks typically produce well-calibrated probabilities on binary classification tasks. While neural networks today are undoubtedly more accurate than they were a decade ago,
we discover with great surprise that \emph{modern neural networks are no longer well-calibrated}.
This is visualized in \autoref{figure.complenet}, which compares a 5-layer LeNet (left) \cite{lecun1998gradient} with a 110-layer ResNet (right) \cite{he2015deep} on the CIFAR-100 dataset.
The top row shows the distribution of prediction confidence (i.e. probabilities associated with the predicted label) as histograms.
The average confidence of LeNet closely matches its accuracy, while the average confidence of the ResNet is substantially higher than its accuracy.
This is further illustrated in the bottom row reliability diagrams \cite{degroot1983comparison,niculescu2005predicting}, which show accuracy as a function of confidence. We see that LeNet is well-calibrated, as confidence closely approximates the expected accuracy (i.e. the bars align roughly along the diagonal). On the other hand, the ResNet's accuracy is better, but does not match its confidence.

Our goal is not only to understand why neural networks have become miscalibrated, but also to identify what methods can alleviate this problem.
In this paper, we demonstrate on several computer vision and NLP tasks that neural networks produce confidences that do not represent true probabilities.
Additionally, we offer insight and intuition into network training and architectural trends that may cause miscalibration.
Finally, we compare various post-processing calibration methods on state-of-the-art neural networks, and introduce several extensions of our own. Surprisingly, we find that a single-parameter variant of Platt scaling \cite{platt1999probabilistic} -- which we refer to as \emph{temperature scaling} -- is often the most effective method at obtaining calibrated probabilities. Because this method is straightforward to implement with existing deep learning frameworks, it can be easily adopted in practical settings.

\section{Definitions}
\label{definitions}
\label{sec:definitions}

 \setlength{\abovedisplayskip}{4pt}
 \setlength{\belowdisplayskip}{4pt}
 \setlength{\textfloatsep}{4pt}

\begin{figure*}[t!]
  \centering
  \includegraphics[width=\textwidth]{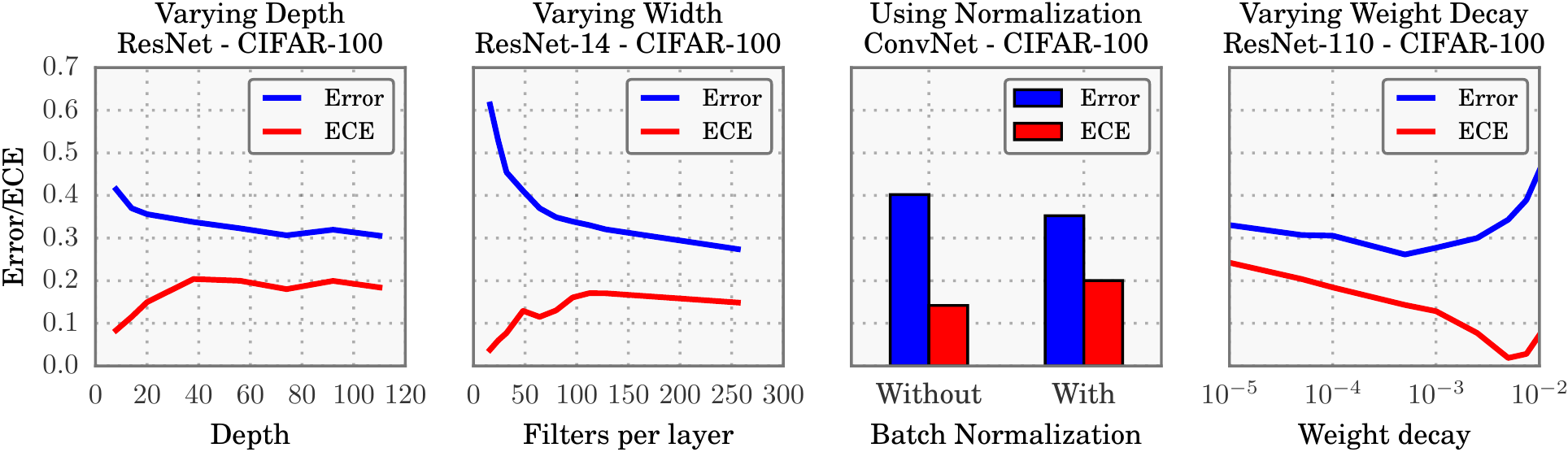}
  \caption{The effect of network depth (far left), width (middle left), Batch Normalization (middle right), and weight decay (far right) on miscalibration, as measured by ECE (lower is better).}
  \label{figure.factors}
  \vspace*{-2ex}
\end{figure*}

The problem we address in this paper is supervised multi-class classification with neural networks. The input $X\in\scrX$ and label $Y\in\scrY=\{1, \ldots, K\}$ are random variables that follow a ground truth joint distribution $\pi(X,Y) = \pi(Y|X) \pi(X)$.
Let $h$ be a neural network with $h(X) = (\yh, \ph)$, where $\yh$ is a class prediction and $\ph$ is its associated confidence, i.e. probability of correctness.
We would like the confidence estimate $\ph$ to be calibrated, which intuitively means that $\ph$ represents a true probability. For example, given 100 predictions, each with confidence of $0.8$, we expect that $80$ should be correctly classified.
More formally, we define \emph{perfect calibration} as
\begin{equation}
\label{perfect_calibration}
\P \left(\yh=Y\sep \ph = p\right)=p,\quad\forall p \in [0,1]
\end{equation}
where the probability is over the joint distribution.
In all practical settings, achieving perfect calibration is impossible.
Additionally, the probability in \eqref{perfect_calibration} cannot be computed using finitely many samples since $\ph$ is a continuous random variable. This motivates the need for empirical approximations that capture the essence of \eqref{perfect_calibration}.

\paragraph{Reliability Diagrams} (e.g. \autoref{figure.complenet} bottom)
are a visual representation of model calibration \citep{degroot1983comparison,niculescu2005predicting}.
These diagrams plot expected sample accuracy as a function of confidence.
If the model is perfectly calibrated -- i.e. if \eqref{perfect_calibration} holds -- then the diagram should plot the identity function. Any deviation from a perfect diagonal represents miscalibration.

To estimate the expected accuracy from finite samples, we group predictions into $M$ interval bins (each of size $1/M$) and calculate the accuracy of each bin.
Let $B_m$ be the set of indices of samples whose prediction confidence falls into the interval $I_m = (\frac{m-1}{M},\frac{m}{M}]$.
The accuracy of $B_m$ is
$$\acc(B_m) = \frac{1}{|B_m|}\sum_{i \in B_m}\ind(\hat{y}_i = y_i),$$
where $\hat{y}_i$ and $y_i$ are the predicted and true class labels for
sample $i$.
Basic probability tells us that $\acc(B_m)$ is an unbiased and consistent estimator of $\P(\yh = Y \mid \ph \in I_m)$. We define the average confidence within bin $B_m$ as
$$\conf(B_m) = \frac{1}{|B_m|}\sum_{i \in B_m}\hat{p}_i,$$
where $\hat{p}_i$ is the confidence for sample $i$.
$\acc(B_m)$ and $\conf(B_m)$ approximate the left-hand and right-hand sides of \eqref{perfect_calibration} respectively for bin $B_m$.
Therefore, a perfectly calibrated model will have $\acc(B_m) = \conf(B_m)$ for all $m \in \{ 1, \ldots, M \}$.
Note that reliability diagrams do not display the proportion of samples in a given bin, and thus cannot be used to estimate how many samples are calibrated.

\paragraph{Expected Calibration Error (ECE).}
While reliability diagrams are useful visual tools, it is more convenient to have a scalar summary statistic of calibration.
Since statistics comparing two distributions cannot be comprehensive, previous works have proposed variants, each with a unique emphasis.
One notion of miscalibration is the difference in expectation between confidence and accuracy, i.e.
\begin{align}
  \E_{\ph} \left[ \left| \P\left(\yh = Y \sep \ph = p\right) - p \right| \right]
  \label{eqn:miscalibration}
\end{align}
Expected Calibration Error \citep{naeini2015obtaining} -- or ECE -- approximates \eqref{eqn:miscalibration} by partitioning predictions into $M$ equally-spaced bins (similar to the reliability diagrams) and taking a weighted average of the bins' accuracy/confidence difference. More precisely,
\begin{equation}
\label{ece}
\text{ECE} = \sum_{m=1}^{M}\frac{|B_m|}{n}\bigg|\acc(B_m) - \conf(B_m)\bigg|,
\end{equation}
where $n$ is the number of samples.
The difference between $\acc$ and $\conf$ for a given bin represents the calibration \emph{gap} (red bars in reliability diagrams -- e.g. \autoref{figure.complenet}).
We use ECE as the primary empirical metric to measure calibration.
See \autoref{sup:definitions} for more analysis of this metric.


\paragraph{Maximum Calibration Error (MCE).} In high-risk applications where reliable confidence measures are absolutely necessary, we may wish to minimize the worst-case deviation between confidence and accuracy:
\begin{equation}
  \max_{p \in [0, 1]} \left| \P\left(\yh = Y \sep \ph = p\right) - p \right|.
\end{equation}
The Maximum Calibration Error \citep{naeini2015obtaining} -- or MCE -- estimates this deviation. Similarly to ECE, this approximation involves binning:
\begin{equation}
\label{mce}
\text{MCE} = \max_{m \in \{1,\ldots,M\} }\left|\acc(B_m) - \conf(B_m)\right|.
\end{equation}
We can visualize MCE and ECE on reliability diagrams.
MCE is the largest calibration gap (red bars) across all bins, whereas ECE is a weighted average of all gaps.
For perfectly calibrated classifiers, MCE and ECE both equal 0.

\paragraph{Negative log likelihood} is a standard measure of a probabilistic model's quality \cite{friedman2001elements}. It is also referred to as the cross entropy loss in the context of deep learning \cite{bengio2015deep}. Given a probabilistic model $\hat{\pi}(Y|X)$ and $n$ samples, NLL is defined as:
\begin{align}
  \scrL = -\sum_{i=1}^{n}\log(\hat{\pi}(y_i|\xb_i))
\end{align}
It is a standard result \cite{friedman2001elements} that, in expectation, NLL is minimized if and only if $\hat{\pi}(Y|X)$ recovers the ground truth conditional distribution $\pi(Y|X)$.

\section{Observing Miscalibration}
\label{motivation}
\setlength{\abovedisplayskip}{4pt}
\setlength{\belowdisplayskip}{4pt}
\setlength{\textfloatsep}{4pt}

The architecture and training procedures of neural networks have rapidly evolved in recent years. In this section we identify some recent changes that are responsible for the miscalibration phenomenon observed in \autoref{figure.complenet}.
Though we cannot claim causality, we find that increased model capacity and lack of regularization are closely related to model miscalibration.



\paragraph{Model capacity.}
The model capacity of neural networks has increased at a dramatic pace over the past few years. It is now common to see networks with hundreds, if not thousands of layers \cite{he2015deep,huang2016deep} and hundreds of convolutional filters per layer \cite{zagoruyko2016wide}. Recent work shows that very deep or wide models are able to generalize better than smaller ones, while exhibiting the capacity to easily fit the training set \cite{zhang2016understanding}.

Although increasing depth and width may reduce classification error, we observe that these increases negatively affect model calibration.
\autoref{figure.factors} displays error and ECE as a function of depth and width on a ResNet trained on CIFAR-100. The far left figure varies depth for a network with 64 convolutional filters per layer, while the middle left figure fixes the depth at 14 layers and varies the number of convolutional filters per layer. Though even the smallest models in the graph exhibit some degree of miscalibration, the ECE metric grows substantially with model capacity.
During training, after the model is able to correctly classify (almost) all training samples, NLL can be further minimized by increasing the confidence of predictions.
Increased model capacity will lower training NLL, and thus the model will be more (over)confident on average.

\paragraph{Batch Normalization} \cite{ioffe2015batch} improves the optimization of neural networks by minimizing distribution shifts in activations within the neural network's hidden layers. Recent research suggests that these normalization techniques have enabled the development of very deep architectures, such as ResNets \cite{he2015deep} and DenseNets \cite{huang2016densely}. It has been shown that Batch Normalization improves training time, reduces the need for additional regularization, and can in some cases improve the accuracy of networks.

While it is difficult to pinpoint exactly how Batch Normalization affects the final predictions of a model, we do observe that models trained with Batch Normalization tend to be more miscalibrated. In the middle right plot of \autoref{figure.factors}, we see that a 6-layer ConvNet obtains worse calibration when Batch Normalization is applied, even though classification accuracy improves slightly. We find that this result holds regardless of the hyperparameters used on the Batch Normalization model (i.e. low or high learning rate, etc.).

\begin{figure}[t!]
	\centering
	\includegraphics[width=\columnwidth]{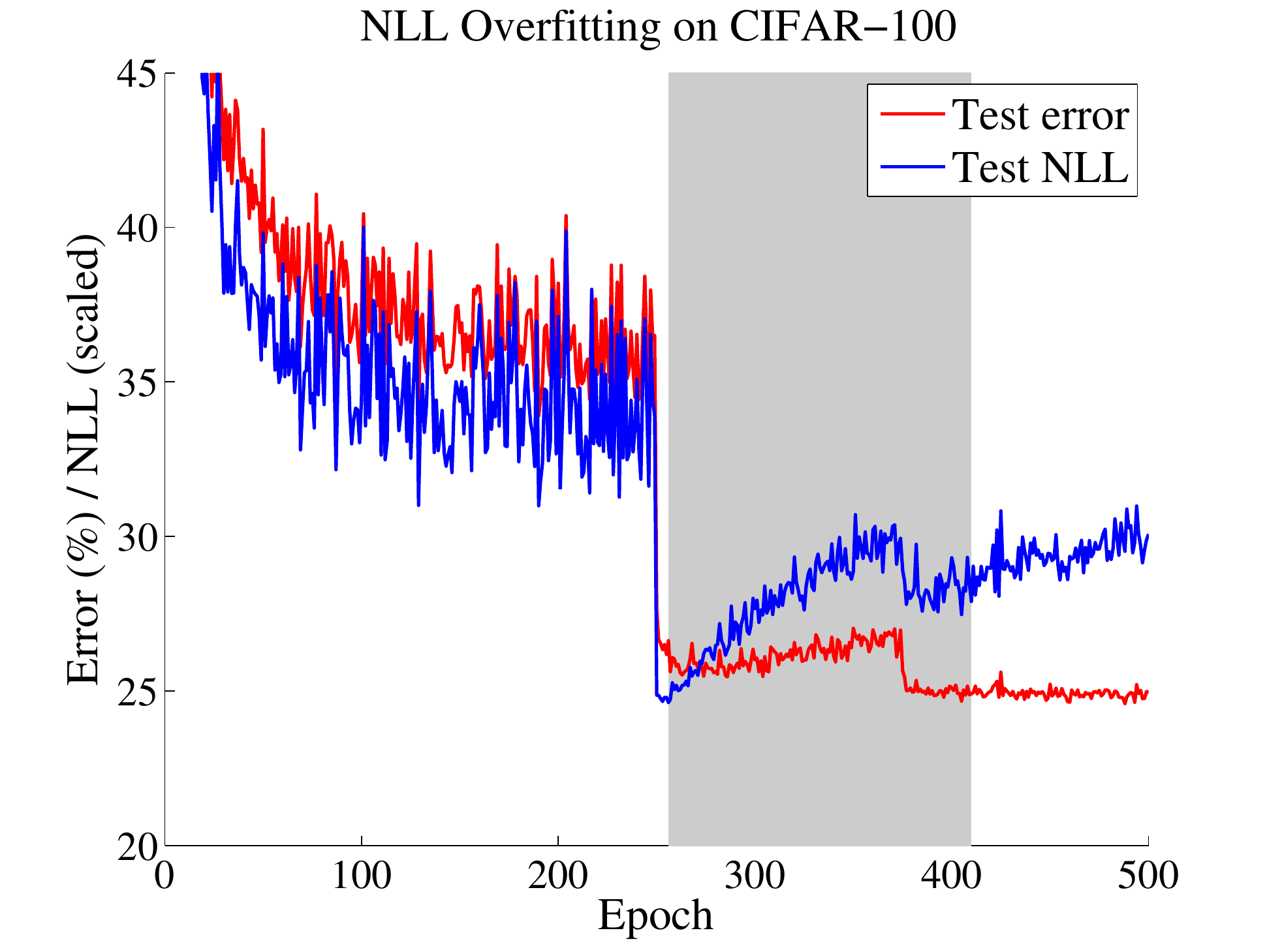}
	\caption{Test error and NLL of a 110-layer ResNet with stochastic depth on CIFAR-100 during training. NLL is scaled by a constant to fit in the figure. Learning rate drops by 10x at epochs 250 and 375. The shaded area marks between epochs at which the best validation \emph{loss} and best validation \emph{error} are produced.}
	\label{figure.cifar100_overfit}
	\vspace{1ex}
\end{figure}

\paragraph{Weight decay,} which used to be the predominant regularization mechanism for neural networks, is decreasingly utilized when training modern neural networks. Learning theory suggests that regularization is necessary to prevent overfitting, especially as model capacity increases \cite{vapnik1998}. However, due to the apparent regularization effects of Batch Normalization, recent research seems to suggest that models with less L2 regularization tend to generalize better \cite{ioffe2015batch}. As a result, it is now common to train models with little weight decay, if any at all. The top performing ImageNet models of 2015 all use an order of magnitude less weight decay than models of previous years \cite{he2015deep, simonyan2014very}.

We find that training with less weight decay has a negative impact on calibration.
The far right plot in \autoref{figure.factors} displays training error and ECE for a 110-layer ResNet with varying amounts of weight decay.
The only other forms of regularization are data augmentation and Batch Normalization.
We observe that calibration and accuracy are not optimized by the same parameter setting.
While the model exhibits both over-regularization and under-regularization with respect to classification error, it does not appear that calibration is negatively impacted by having too much weight decay.
Model calibration continues to improve when more regularization is added, well after the point of achieving optimal accuracy.
The slight uptick at the end of the graph may be an artifact of using a weight decay factor that impedes optimization.

\paragraph{NLL} can be used to indirectly measure model calibration. In practice, we observe \emph{a disconnect between NLL and accuracy}, which may explain the miscalibration in \autoref{figure.factors}.
This disconnect occurs because neural networks can \emph{overfit to NLL without overfitting to the 0/1 loss}.
We observe this trend in the training curves of some miscalibrated models.
\autoref{figure.cifar100_overfit} shows test error and NLL (rescaled to match error) on CIFAR-100 as training progresses.
Both error and NLL immediately drop at epoch 250, when the learning rate is dropped; however, NLL overfits during the remainder of training.
Surprisingly, overfitting to NLL is beneficial to classification accuracy. On CIFAR-100, test error drops from $29\%$ to $27\%$ in the region where NLL overfits. This phenomenon renders a concrete explanation of miscalibration: the network learns better classification accuracy at the expense of well-modeled probabilities.

We can connect this finding to recent work examining the generalization of large neural networks. \citet{zhang2016understanding} observe that deep neural networks seemingly violate  the common understanding of learning theory that large models with little regularization will not generalize well. The observed disconnect between NLL and 0/1 loss suggests that these high capacity models are not necessarily immune from overfitting, but rather, overfitting manifests in probabilistic error rather than classification error.

\section{Calibration Methods}
\label{related}
\setlength{\abovedisplayskip}{4pt}
\setlength{\belowdisplayskip}{4pt}
\setlength{\textfloatsep}{4pt}

In this section, we first review existing calibration methods, and introduce new variants of our own.
All methods are post-processing steps that produce (calibrated) probabilities.
Each method requires a hold-out validation set, which in practice can be the same set used for hyperparameter tuning.
We assume that the training, validation, and test sets are drawn from the same distribution.

\subsection{Calibrating Binary Models}
\label{binary_calibration}

We first introduce calibration in the binary setting, i.e. $\mathcal{Y}=\{0,1\}$.
For simplicity, throughout this subsection, we assume the model outputs only the confidence for the positive class.\footnote{
  This is in contrast with the setting in \autoref{definitions}, in which the model produces both a class prediction and confidence.
}
Given a sample $\xb_i$, we have access to $\hat p_i$ -- the network's predicted probability of $y_i = 1$, as well as $z_i \in \mathbb{R}$ -- which is the network's non-probabilistic output, or \emph{logit}. The predicted probability $\hat p_i$ is derived from $z_i$ using a sigmoid function $\sigma$; i.e.
$\hat p_i = \sigma(z_i)$. Our goal is to produce a calibrated probability $\hat q_i$ based on $y_i$, $\hat p_i$, and $z_i$.

\paragraph{Histogram binning} \cite{zadrozny2001obtaining} is a simple non-parametric calibration method.
In a nutshell,  all uncalibrated predictions $\hat{p}_i$ are divided
into mutually exclusive bins $B_1,\dots,B_M$. Each bin is assigned a calibrated score $\theta_m$; i.e. if $\hat p_i$ is assigned to bin $B_m$, then $\hat q_i = \theta_m$. At test time, if prediction $\hat{p}_{te}$ falls into bin $B_m$, then the calibrated prediction $\hat q_{te}$ is $\theta_m$.
More precisely, for a suitably chosen $M$ (usually small), we first define bin boundaries $0 = a_1 \leq a_2 \leq \ldots \leq a_{M+1} = 1$, where the bin $B_m$ is defined by the interval $(a_m, a_{m+1}]$.
Typically the bin boundaries are either chosen to be equal length intervals
or to equalize the number of samples in each bin.
The predictions $\theta_i$ are chosen to minimize the bin-wise squared loss:
\begin{equation}
\min_{\theta_1,\ldots,\theta_M} \: \sum_{m=1}^M \sum_{i = 1}^n
\mathbf{1} (a_m \leq \hat p_i < a_{m+1}) \left(\theta_m - y_i \right)^2,
\label{eqn:hist_bin}
\end{equation}
where $\mathbf{1}$ is the indicator function. Given fixed bins boundaries, the solution to \eqref{eqn:hist_bin} results in $\theta_m$ that correspond to the average number of positive-class samples in bin $B_m$.

\paragraph{Isotonic regression} \cite{zadrozny2002transforming}, arguably the most common non-parametric calibration method,
learns a piecewise constant function $f$ to transform uncalibrated outputs; i.e. $\hat q_i = f(\hat p_i)$.
Specifically, isotonic regression produces $f$ to minimize the square loss $\sum_{i=1}^n (f(\hat p_i) - y_i)^2$.
Because $f$ is constrained to be piecewise constant, we can write the optimization problem as:
\begin{equation}
\begin{aligned}
\label{iso_eq}
\min_{\substack{M \\ \theta_1,\ldots,\theta_M \\ a_1,\ldots,a_{M+1}}} & \hspace{3pt} \sum_{m=1}^M \sum_{i = 1}^n
\mathbf{1} (a_m \leq \hat p_i < a_{m+1}) \left(\theta_m - y_i \right)^2 \\
\text{subject to} & \hspace{8pt} 0 = a_1 \leq a_2 \leq \ldots \leq a_{M+1} = 1, \nonumber \\
& \hspace{8pt} \theta_1 \leq \theta_2 \leq \ldots \leq \theta_M. \nonumber
\end{aligned}
\end{equation}
where $M$ is the number of intervals; $a_1, \ldots, a_{M+1}$ are the interval boundaries; and $\theta_1, \ldots, \theta_M$ are the function values.
Under this parameterization, isotonic regression is a strict generalization of histogram binning in which the bin boundaries and bin predictions are jointly optimized.

%
%

\paragraph{Bayesian Binning into Quantiles (BBQ)} \cite{naeini2015obtaining} is a extension of histogram binning using Bayesian model averaging. Essentially, BBQ marginalizes out all possible \emph{binning schemes} to produce $\hat q_i$.
More formally, a binning scheme $s$ is a pair $(M,\mathcal{I})$ where $M$ is the number of bins, and $\mathcal{I}$ is a corresponding partitioning of $[0,1]$ into disjoint intervals ($0 = a_1 \leq a_2 \leq \ldots \leq a_{M+1} = 1$). The parameters of a binning scheme are $\theta_1,\ldots,\theta_M$. Under this framework, histogram binning and isotonic regression both produce a single binning scheme, whereas BBQ considers a space $\mathcal{S}$ of all possible binning schemes for the validation dataset $D$. BBQ performs Bayesian averaging of the probabilities produced by each scheme:\footnote{
  Because the validation dataset is finite, $\mathcal{S}$ is as well.
}
%
%
\begin{align*}
\P(\hat q_{te} \sep \hat p_{te}, D) &= \sum_{s \in \mathcal{S}} \P(\hat q_{te}, S=s \sep \hat p_{te}, D) \\
  &= \sum_{s \in \mathcal{S}} \P(\hat q_{te} \sep \hat p_{te}, S\!=\!s, D) \P(S\!=\!s \sep D).
\end{align*}
where $\P(\hat q_{te} \sep \hat p_{te}, S\!=\!s, D)$ is the calibrated probability using binning scheme $s$. Using a uniform prior, the weight $\P(S\!=\!s \sep D)$ can be derived using Bayes' rule:
$$\P(S\!=\!s \sep D) = \frac{\P(D \sep S\!=\!s)}{\sum_{s' \in \mathcal{S}} \P(D \sep S\!=\!s')}.$$
The parameters $\theta_1,\ldots,\theta_M$ can be viewed as parameters of $M$ independent binomial distributions. Hence, by placing a Beta prior on $\theta_1, \ldots, \theta_M$, we can obtain a closed form expression for the marginal likelihood $\P(D \sep S\!=\!s)$. This allows us to compute $\P(\hat q_{te} \sep \hat p_{te}, D)$ for any test input.

\paragraph{Platt scaling} \cite{platt1999probabilistic} is a parametric approach to calibration, unlike the other approaches. The non-probabilistic predictions of a classifier are used as features for a logistic regression model, which is trained on the validation set to return probabilities. More specifically, in the context of neural networks \cite{niculescu2005predicting}, Platt scaling learns scalar parameters $a,b \in \mathbb{R}$ and outputs $\hat q_i = \sigma(a z_i +b)$ as the calibrated probability. Parameters $a$ and $b$ can be optimized using the NLL loss over the validation set. It is important to note that the neural network's parameters are fixed during this stage.


\begin{table*}[t!]
	\centering
	\resizebox{\textwidth}{!}{%
\begin{tabular}{c c c c c c c c c}
\toprule
          Dataset &            Model & Uncalibrated & Hist. Binning & Isotonic &      BBQ & Temp. Scaling & Vector Scaling & Matrix Scaling \\
\midrule
            Birds &        ResNet 50 &        9.19\% &         4.34\% &    5.22\% &    4.12\% &       {\bf1.85\%} &           3.0\% &         21.13\% \\
             Cars &        ResNet 50 &         4.3\% &       {\bf1.74\%} &    4.29\% &    1.84\% &         2.35\% &          2.37\% &          10.5\% \\
         CIFAR-10 &       ResNet 110 &         4.6\% &         0.58\% &    0.81\% &  {\bf0.54\%} &         0.83\% &          0.88\% &           1.0\% \\
         CIFAR-10 &  ResNet 110 (SD) &        4.12\% &         0.67\% &    1.11\% &     0.9\% &        {\bf0.6\%} &          0.64\% &          0.72\% \\
         CIFAR-10 &   Wide ResNet 32 &        4.52\% &         0.72\% &    1.08\% &    0.74\% &       {\bf0.54\%} &           0.6\% &          0.72\% \\
         CIFAR-10 &      DenseNet 40 &        3.28\% &         0.44\% &    0.61\% &    0.81\% &       {\bf0.33\%} &          0.41\% &          0.41\% \\
         CIFAR-10 &          LeNet 5 &        3.02\% &         1.56\% &    1.85\% &    1.59\% &       {\bf0.93\%} &          1.15\% &          1.16\% \\
        CIFAR-100 &       ResNet 110 &       16.53\% &         2.66\% &    4.99\% &    5.46\% &       {\bf1.26\%} &          1.32\% &         25.49\% \\
        CIFAR-100 &  ResNet 110 (SD) &       12.67\% &         2.46\% &    4.16\% &    3.58\% &         0.96\% &         {\bf0.9\%} &         20.09\% \\
        CIFAR-100 &   Wide ResNet 32 &        15.0\% &         3.01\% &    5.85\% &    5.77\% &       {\bf2.32\%} &          2.57\% &         24.44\% \\
        CIFAR-100 &      DenseNet 40 &       10.37\% &         2.68\% &    4.51\% &    3.59\% &         1.18\% &        {\bf1.09\%} &         21.87\% \\
        CIFAR-100 &          LeNet 5 &        4.85\% &         6.48\% &    2.35\% &    3.77\% &       {\bf2.02\%} &          2.09\% &         13.24\% \\
         ImageNet &     DenseNet 161 &        6.28\% &         4.52\% &    5.18\% &    3.51\% &       {\bf1.99\%} &          2.24\% &              - \\
         ImageNet &       ResNet 152 &        5.48\% &         4.36\% &    4.77\% &    3.56\% &       {\bf1.86\%} &          2.23\% &              - \\
             SVHN &  ResNet 152 (SD) &        0.44\% &       {\bf0.14\%} &    0.28\% &    0.22\% &         0.17\% &          0.27\% &          0.17\% \\
              \midrule
          20 News &            DAN 3 &        8.02\% &        {\bf3.6\%} &    5.52\% &    4.98\% &         4.11\% &          4.61\% &           9.1\% \\
          Reuters &            DAN 3 &        0.85\% &         1.75\% &    1.15\% &    0.97\% &         0.91\% &        {\bf0.66\%} &          1.58\% \\
       SST Binary &         TreeLSTM &        6.63\% &         1.93\% &  {\bf1.65\%} &    2.27\% &         1.84\% &          1.84\% &          1.84\% \\
 SST Fine Grained &         TreeLSTM &        6.71\% &         2.09\% &  {\bf1.65\%} &    2.61\% &         2.56\% &          2.98\% &          2.39\% \\
\bottomrule
\end{tabular}
}

  \caption{ECE (\%) (with $M=15$ bins) on standard vision and NLP datasets before calibration and with various calibration methods. The number following a model's name denotes the network depth.}
	\label{table.ece}
	\vspace{-2ex}
\end{table*}

\subsection{Extension to Multiclass Models}

For classification problems involving $K>2$ classes, we return to the original problem formulation.
The network outputs a class prediction $\hat y_i$ and confidence score $\hat p_i$ for each input $\xb_i$. In this case, the network logits $\zb_i$ are vectors, where $\hat y_i = \argmax_{k} z_i^{(k)}$, and $\hat p_i$ is typically derived using the softmax function $\sigma_\text{SM}$:
$$\sigma_\text{SM}(\zb_i)^{(k)} = \frac{\exp(z_i^{(k)})}
{\sum_{j=1}^K \exp(z_i^{(j)})}, \hspace{10pt}
\hat p_i = \max_k \: \sigma_\text{SM}(\zb_i)^{(k)}.$$
%
%
The goal is to produce a calibrated confidence $\hat q_i$ and (possibly new) class prediction $\hat y_i'$
based on $y_i$, $\hat y_i$, $\hat p_i$, and $\zb_i$.


\newcommand{\bb}{\mathbf{b}}
\newcommand{\Wb}{\mathbf{W}}

\paragraph{Extension of binning methods.} One common way of extending binary calibration methods to the multiclass setting is by treating the problem as $K$ one-versus-all problems \cite{zadrozny2002transforming}.
For $k = 1,\ldots,K$, we form a binary calibration problem where the label is $\ind(y_i = k)$ and the predicted probability is $\sigma_\text{SM}(\zb_i)^{(k)}$. This gives us $K$ calibration models, each for a particular class. At test time, we obtain an unnormalized probability vector $[ \hat q_i^{(1)}, \ldots, \hat q_i^{(K)} ]$, where $\hat q_i^{(k)}$ is the calibrated probability for class $k$. The new class prediction $\hat y_i'$ is the argmax of the vector, and the new confidence $\hat q_i'$ is the max of the vector normalized by $\sum_{k=1}^K \hat q_i^{(k)}$. This extension can be applied to histogram binning, isotonic regression, and BBQ.

\paragraph{Matrix and vector scaling} are two multi-class extensions of Platt scaling. Let $\zb_i$ be the \emph{logits vector} produced before the softmax layer for input $\xb_i$. \emph{Matrix scaling applies} a linear transformation $\Wb \zb_i + \bb$ to the logits:
\begin{equation}
\label{scaling}
\begin{aligned}
\hat q_i  &= \max_k \: \sigma_\text{SM} ( \Wb \zb_i + \bb)^{(k)}, \\
\hat y_i' &= \argmax_k \: (\Wb \zb_i + \bb)^{(k)}.
\end{aligned}
\end{equation}
The parameters $\Wb$ and $\bb$ are optimized with respect to NLL on the validation set. As the number of parameters for matrix scaling grows quadratically with the number of classes $K$, we define \emph{vector scaling} as a variant where $\Wb$ is restricted to be a diagonal matrix.

\paragraph{Temperature scaling,} the simplest extension of Platt scaling, uses a single scalar parameter $T > 0$ for all classes. Given the logit vector $\zb_i$, the new confidence prediction is
\begin{equation}
\label{temp}
\hat q_i = \max_k \: \sigma_\text{SM} ( \zb_i / T)^{(k)}.
\end{equation}
$T$ is called the temperature, and it ``softens'' the softmax (i.e. raises the output entropy) with $T > 1$.
As $T \rightarrow \infty$, the probability $\hat q_i$ approaches $1/K$, which represents maximum uncertainty. With $T=1$, we recover the original probability $\hat p_i$.
As $T \rightarrow 0$, the probability collapses to a point mass (i.e. $\hat q_i = 1$).
$T$ is optimized with respect to NLL on the validation set.
Because the parameter $T$ does not change the maximum of the softmax function, the class prediction $\hat y_i'$ remains unchanged. In other words, \emph{temperature scaling does not affect the model's accuracy.}

Temperature scaling is commonly used in settings such as knowledge distillation \citep{hinton2015distilling} and statistical mechanics \citep{jaynes1957information}. To the best of our knowledge, we are not aware of any prior use in the context of calibrating probabilistic models.\footnote{To highlight the connection with prior works we define temperature scaling in terms of $\frac{1}{T}$ instead of a multiplicative scalar.}
The model is equivalent to maximizing the entropy of the output probability distribution subject to certain constraints on the logits (see \autoref{sup:proof}).


\subsection{Other Related Works}
Calibration and confidence scores have been studied in various contexts in recent years. \citet{kuleshovE16} study the problem of calibration in the online setting, where the inputs can come from a potentially adversarial source. \citet{kuleshov2015calibrated} investigate how to produce calibrated probabilities when the output space is a structured object.
\citet{lakshminarayanan2016simple} use ensembles of networks to obtain uncertainty estimates.
 \citet{pereyra2017regularizing} penalize overconfident predictions as a form of regularization.
\citet{hendrycks2016baseline} use confidence scores to determine if samples are out-of-distribution.

Bayesian neural networks \cite{denker1990transforming,mackay1992practical} return a probability distribution over outputs as an alternative way to represent model uncertainty.
\citet{gal2015dropout} draw a connection between Dropout \cite{srivastava2014dropout} and model uncertainty, claiming that sampling models with dropped nodes is a way to estimate the probability distribution over all possible models for a given sample.
\citet{kendall2017uncertainties} combine this approach with a model that outputs a predictive mean and variance for each data point.
This notion of uncertainty is not restricted to classification problems.
Additionally, neural networks can be used in conjunction with Bayesian models that output complete distributions.
For example, deep kernel learning \cite{wilson2016stochastic,wilson2016deep,al2016learning} combines deep neural networks with Gaussian processes on classification and regression problems.
In contrast, our framework, which does not augment the neural network model, returns a confidence score rather than returning a distribution of possible outputs.

\section{Results}
\label{results}
 \setlength{\abovedisplayskip}{4pt}
 \setlength{\belowdisplayskip}{4pt}
 \setlength{\textfloatsep}{4pt}

We apply the calibration methods in \autoref{related} to image classification and document classification neural networks.
For image classification we use 6 datasets:
\begin{enumerate}[noitemsep,nolistsep]
\item Caltech-UCSD Birds \cite{cubdataset}: 200 bird species. 5994/2897/2897 images for train/validation/test sets.
\item Stanford Cars \cite{carsdataset}: 196 classes of cars by make, model, and year. 8041/4020/4020 images for train/validation/test.
\item ImageNet 2012 \cite{deng2009imagenet}: Natural scene images from 1000 classes. 1.3 million/25,000/25,000 images for train/validation/test.
\item CIFAR-10/CIFAR-100 \cite{krizhevsky2009learning}: Color images ($32\times 32$) from 10/100 classes. 45,000/5,000/10,000 images for train/validation/test.
\item Street View House Numbers (SVHN) \cite{netzer2011reading}: $32\times 32$ colored images of cropped out house numbers from Google Street View.
598,388/6,000/26,032 images for train/validation/test.
\end{enumerate}
We train state-of-the-art convolutional networks: ResNets \cite{he2015deep}, ResNets with stochastic depth (SD) \cite{huang2016deep}, Wide ResNets \cite{zagoruyko2016wide}, and DenseNets \cite{huang2016densely}. We use the data preprocessing, training procedures, and hyperparameters as described in each paper. For Birds and Cars, we fine-tune networks pretrained on ImageNet.

For document classification we experiment with 4 datasets:
\begin{enumerate}[noitemsep,nolistsep]
\item 20 News: News articles, partitioned into 20 categories by content. 9034/2259/7528 documents for train/validation/test.
\item Reuters: News articles, partitioned into 8 categories by topic. 4388/1097/2189 documents for train/validation/test.
\item Stanford Sentiment Treebank (SST) \cite{socher2013recursive}: Movie reviews, represented as sentence parse trees that are annotated by sentiment. Each sample includes a coarse binary label and a fine grained 5-class label.
As described in \cite{tai2015improved}, the training/validation/test sets contain 6920/872/1821 documents for binary, and 544/1101/2210 for fine-grained.
\end{enumerate}
On 20 News and Reuters, we train Deep Averaging Networks (DANs) \cite{iyyer2015deep} with 3 feed-forward layers and Batch Normalization.  On SST, we train \mbox{TreeLSTMs} (Long Short Term Memory) \cite{tai2015improved}.
For both models we use the default hyperparmaeters suggested by the authors.

\begin{figure*}[h!]
	\centering
	\includegraphics[width=0.98\textwidth]{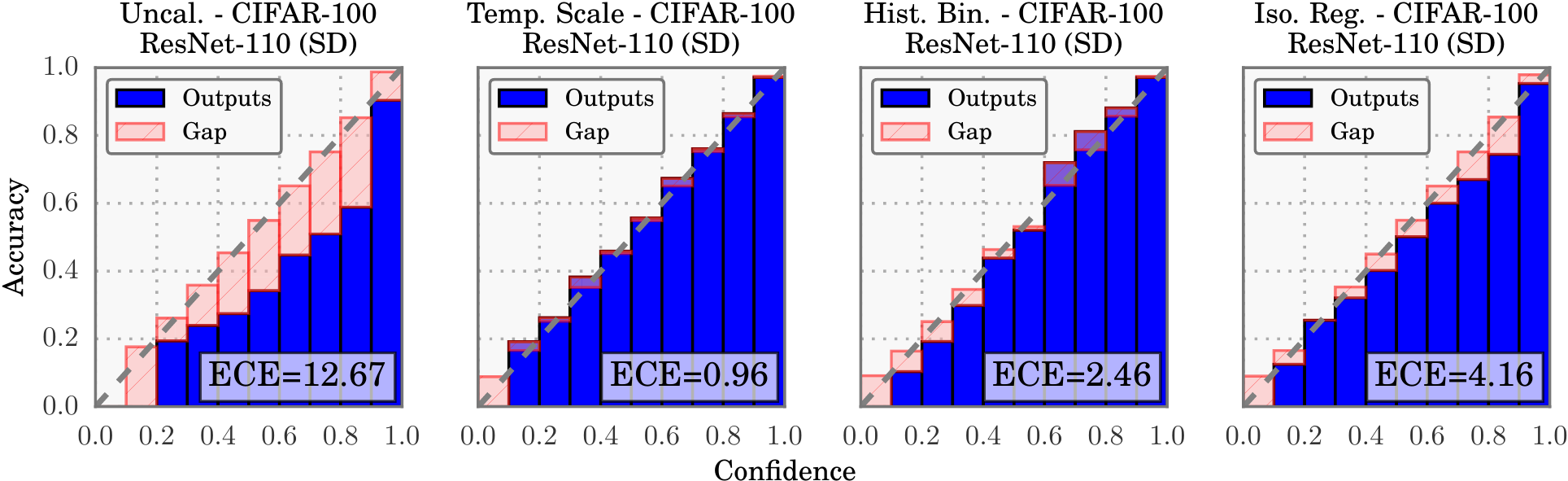}
	\caption{Reliability diagrams for CIFAR-100 before (far left) and after calibration (middle left, middle right, far right).}
	\label{figure.reliability}
\end{figure*}

\paragraph{Calibration Results.}
\autoref{table.ece} displays model calibration, as measured by ECE (with $M=15$ bins), before and after applying the various methods (see \autoref{sup:tables} for MCE, NLL, and error tables).
It is worth noting that most datasets and models experience some degree of miscalibration, with ECE typically between $4$ to $10\%$.
This is not architecture specific: we observe miscalibration on convolutional networks (with and without skip connections), recurrent networks, and deep averaging networks.
The two notable exceptions are SVHN and Reuters, both of which experience ECE values below $1\%$. Both of these datasets have very low error ($1.98\%$ and $2.97\%$, respectively); and therefore the ratio of ECE to error is comparable to other datasets.

Our most important discovery is the \emph{surprising effectiveness of temperature scaling} despite its remarkable simplicity. Temperature scaling outperforms all other methods on the vision tasks, and performs comparably to other methods on the NLP datasets. What is perhaps even more surprising is that temperature scaling outperforms the vector and matrix Platt scaling variants, which are strictly more general methods. In fact, vector scaling recovers essentially the same solution as temperature scaling -- the learned vector has nearly constant entries, and therefore is no different than a scalar transformation. In other words, network miscalibration is intrinsically low dimensional.

The only dataset that temperature scaling does not calibrate is the Reuters dataset. In this instance, only one of the above methods is able to improve calibration. Because this dataset is well-calibrated to begin with (ECE $\leq 1\%$), there is not much room for improvement with any method, and post-processing may not even be necessary to begin with. It is also possible that our measurements are affected by dataset split or by the particular binning scheme.


Matrix scaling performs poorly on datasets with hundreds of classes (i.e. Birds, Cars, and CIFAR-100), and fails to converge on the 1000-class ImageNet dataset.
This is expected, since the number of parameters scales quadratically with the number of classes.
Any calibration model with tens of thousands (or more) parameters will overfit to a small validation set, even when applying regularization.

Binning methods improve calibration on most datasets, but do not outperform temperature scaling. Additionally, binning methods tend to change class predictions which hurts accuracy (see \autoref{sup:tables}). Histogram binning, the simplest binning method, typically outperforms isotonic regression and BBQ, despite the fact that both methods are strictly more general. This further supports our finding that calibration is best corrected by simple models.

\paragraph{Reliability diagrams.} \autoref{figure.reliability} contains reliability diagrams for 110-layer ResNets on CIFAR-100 before and after calibration. From the far left diagram, we see that the uncalibrated ResNet tends to be overconfident in its predictions. We then can observe the effects of temperature scaling (middle left), histogram binning (middle right), and isotonic regression (far right) on calibration. All three displayed methods produce much better confidence estimates. Of the three methods, temperature scaling most closely recovers the desired diagonal function. Each of the bins are well calibrated, which is remarkable given that all the probabilities were modified by only a single parameter. We include reliability diagrams for other datasets in \autoref{sup:reliability}.

\paragraph{Computation time.} All methods scale linearly with the number of validation set samples. Temperature scaling is by far the fastest method, as it amounts to a one-dimensional convex optimization problem. Using a conjugate gradient solver, the optimal temperature can be found in 10 iterations, or a fraction of a second on most modern hardware. In fact, even a naive line-search for the optimal temperature is faster than any of the other methods. The computational complexity of vector and matrix scaling are linear and quadratic respectively in the number of classes, reflecting the number of parameters in each method. For CIFAR-100 ($K=100$), finding a near-optimal vector scaling solution with conjugate gradient descent requires at least 2 orders of magnitude more time. Histogram binning and isotonic regression take an order of magnitude longer than temperature scaling, and BBQ takes roughly 3 orders of magnitude more time.

\paragraph{Ease of implementation.} BBQ is arguably the most difficult to implement, as it requires implementing a model averaging scheme.
While all other methods are relatively easy to implement, temperature scaling may arguably be the most straightforward to incorporate into a neural network pipeline.
In Torch7 \citep{collobert2011torch7}, for example, we implement temperature scaling by inserting a \texttt{nn.MulConstant} between the logits and the softmax, whose parameter is $1/T$.
We set $T\!=\!1$ during training, and subsequently find its optimal value on the validation set.\footnote{
  For an example implementation, see \url{http://github.com/gpleiss/temperature_scaling}.
}

\section{Conclusion}
\label{conclusion}


Modern neural networks exhibit a strange phenomenon: probabilistic error and miscalibration worsen even as classification error is reduced.
We have demonstrated that recent advances in neural network architecture and training -- model capacity, normalization, and regularization -- have strong effects on network calibration.
It remains future work to understand why these trends affect calibration while improving accuracy.
Nevertheless, simple techniques can effectively remedy the miscalibration phenomenon in neural networks.
Temperature scaling is the simplest, fastest, and most straightforward of the methods, and surprisingly is often the most effective.

\section*{Acknowledgments}
The authors are supported in part by the III-1618134, III-1526012, and IIS-1149882 grants from the
National Science Foundation, as well as the Bill and
Melinda Gates Foundation and the Office of Naval Research.



\bibliography{citations}
\bibliographystyle{icml2017}

%
%

\renewcommand{\thesection}{S\arabic{section}}
\renewcommand{\thesubsection}{\thesection.\arabic{subsection}}

\newcommand{\beginsupplementary}{%
	\setcounter{table}{0}
	\renewcommand{\thetable}{S\arabic{table}}%
	\setcounter{figure}{0}
	\renewcommand{\thefigure}{S\arabic{figure}}%
	\setcounter{section}{0}
}
\newcommand{\suptitl}{Supplementary Materials for:\\ \titl}
\newcommand{\suptitlrunning}{Supplementary Materials: \titlshort}
\beginsupplementary
\icmltitlerunning{\suptitlrunning}
\twocolumn[
  \icmltitle{\suptitl}
  \vskip 0.3in
]
\section{Further Information on Calibration Metrics}
\label{sup:definitions}

We can connect the ECE metric with our exact miscalibration definition, which is restated here:
\begin{align*}
  \E_{\ph} \left[ \left| \P\left(\yh = Y \sep \ph = p\right) - p \right| \right]
  \label{eqn:miscalibration}
\end{align*}
Let $F_{\ph}(p)$ be the cumulative distribution function of $\ph$ so that $F_{\ph}(b) - F_{\ph}(a) = \P(\ph \in [a,b])$. Using the Riemann-Stieltjes integral we have
\begin{flalign*}
&\E_{\ph} \left[ \left| \P\left(\yh = Y \sep \ph = p\right) - p \right| \right] \\
&= \int_0^1 \left| \P\left(\yh = Y \sep \ph=p\right) - p \right| dF_{\ph}(p) \\
&\approx \sum_{m=1}^{M} \left| \P(\yh=Y|\ph = p_m) - p_m \right| \P(\ph \in I_m)
\end{flalign*}
where $I_m$ represents the interval of bin $B_m$. $\left| \P(\yh=Y|\ph = p_m) - p_m \right|$ is closely approximated by $\left| \text{acc}(B_m)  - \hat{p}(B_m) \right|$ for $n$ large. Hence ECE using $M$ bins converges to the $M$-term Riemann-Stieltjes sum of $\E_{\ph} \left[ \left| \P\left(\yh = Y \sep \ph = p\right) - p \right| \right]$.

\section{Further Information on Temperature Scaling}
\label{sec:proofs}

\label{sup:proof}
Here we derive the temperature scaling model using the entropy maximization principle with an appropriate balanced equation.
{\claim \label{max-entropy} Given $n$ samples' logit vectors $\zb_1, \ldots, \zb_n$ and class labels $y_1, \ldots, y_n$, temperature scaling is the unique solution $q$ to the following entropy maximization problem:
\begin{align*}
\max_{q} \hspace{8pt} & -\sum_{i=1}^n \sum_{k=1}^K q(\zb_i)^{(k)} \log q(\zb_i)^{(k)} \\
\text{subject to} \hspace{10pt} & q(\zb_i)^{(k)} \geq 0 \hspace{30pt} \forall i, k \\
                                & \sum_{k=1}^K q(\zb_i)^{(k)} = 1 \hspace{12pt} \forall i \\
																& \sum_{i=1}^n z_i^{(y_i)} = \sum_{i=1}^n \sum_{k=1}^K z_i^{(k)} q(\zb_i)^{(k)}.
\end{align*}}
The first two constraint ensure that $q$ is a probability distribution, while the last constraint limits the scope of distributions. Intuitively, the constraint specifies that the average true class logit is equal to the average weighted logit.
\begin{proof}
We solve this constrained optimization problem using the Lagrangian. We first ignore the constraint $q(\zb_i)^{(k)}$ and later show that the solution satisfies this condition. Let $\lambda,\beta_1,\ldots,\beta_n \in \mathbb{R}$ be the Lagrangian multipliers and define
\begin{align*}
L = &-\sum_{i=1}^n \sum_{k=1}^K q(\zb_i)^{(k)} \log q(\zb_i)^{(k)} \\
    & + \lambda \sum_{i=1}^n \left[ \sum_{k=1}^K z_i^{(k)} q(\zb_i)^{(k)} - z_i^{(y_i)} \right] \\
	  & + \sum_{i=1}^n \beta_i \sum_{k=1}^K (q(\zb_i)^{(k)} - 1).
\end{align*}
Taking the derivative with respect to $q(\zb_i)^{(k)}$ gives $$\frac{\partial}{\partial q(\zb_i)^{(k)}} L = -nK - \log q(\zb_i)^{(k)} + \lambda z_i^{(k)} + \beta_i.$$
Setting the gradient of the Lagrangian $L$ to 0 and rearranging gives $$q(\zb_i)^{(k)} = e^{\lambda z_i^{(k)} + \beta_i - nK}.$$
Since $\sum_{k=1}^K q(\zb_i)^{(k)} = 1$ for all $i$, we must have $$q(\zb_i)^{(k)} = \frac{e^{\lambda z_i^{(k)}}}{\sum_{j=1}^K e^{\lambda z_i^{(j)}}},$$
which recovers the temperature scaling model by setting $T = \frac{1}{\lambda}$. 
\end{proof}

\begin{figure*}[h!]
	\centering
	\includegraphics[width=1.5\columnwidth]{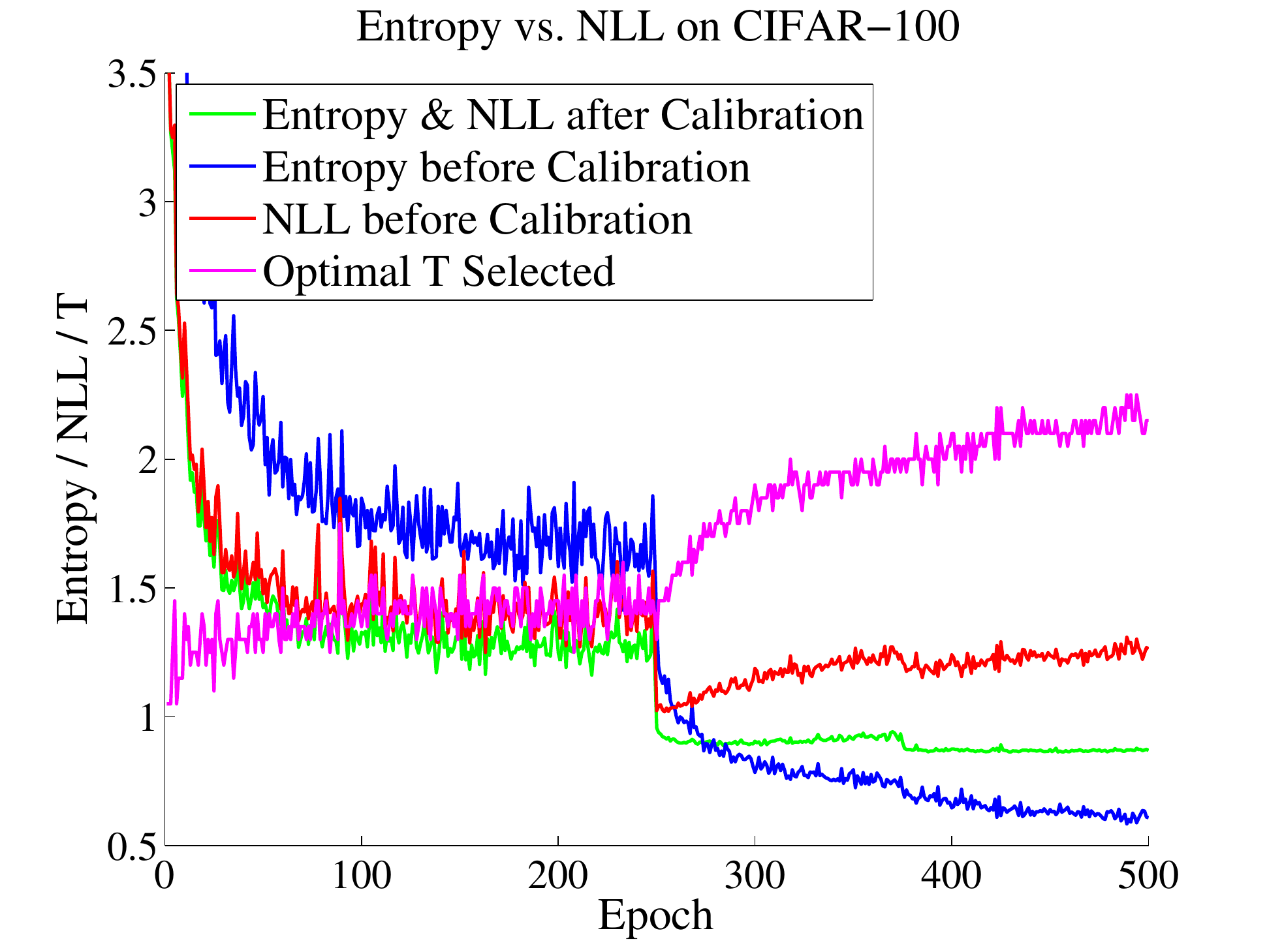}
	\caption{Entropy and NLL for CIFAR-100 before and after calibration. The optimal $T$ selected by temperature scaling rises throughout optimization, as the pre-calibration entropy decreases steadily.
	The post-calibration entropy and NLL on the validation set coincide 
	(which can be derived from the gradient optimality condition of $T$). }
	\label{figure.entropy}
\end{figure*}

\autoref{figure.entropy} visualizes Claim \autoref{max-entropy}.
We see that, as training continues, the model begins to overfit with respect to NLL (red line).
This results in a low-entropy softmax distribution over classes (blue line), which explains the model's overconfidence. Temperature scaling not only lowers the NLL but also raises the entropy of the distribution (green line).

\section{Additional Tables}
\label{sup:tables}

Tables~\ref{table:mce}, \ref{table:err}, and \ref{table:nll} display the MCE, test error, and NLL for all the experimental settings outlined in \autoref{results}.

\begin{table*}[t!]
	\centering
	\resizebox{\textwidth}{!}{%
\begin{tabular}{c c c c c c c c c}
\toprule
          Dataset &            Model & Uncalibrated & Hist. Binning & Isotonic &      BBQ & Temp. Scaling & Vector Scaling & Matrix Scaling \\
\midrule
            Birds &        ResNet 50 &       30.06\% &        25.35\% &   16.59\% &   11.72\% &       {\bf9.08\%} &          9.81\% &         38.67\% \\
             Cars &        ResNet 50 &       41.55\% &       {\bf5.16\%} &   15.23\% &    9.31\% &        20.23\% &          8.59\% &         29.65\% \\
         CIFAR-10 &       ResNet 110 &       33.78\% &        26.87\% &   {\bf7.8\%} &   72.64\% &         8.56\% &         27.39\% &         22.89\% \\
         CIFAR-10 &  ResNet 110 (SD) &       34.52\% &         17.0\% &   16.45\% &   19.26\% &        15.45\% &         15.55\% &       {\bf10.74\%} \\
         CIFAR-10 &   Wide ResNet 32 &       27.97\% &        12.19\% &    6.19\% &    9.22\% &         9.11\% &        {\bf4.43\%} &          9.65\% \\
         CIFAR-10 &      DenseNet 40 &       22.44\% &         7.77\% &   19.54\% &   14.57\% &         4.58\% &        {\bf3.17\%} &          4.36\% \\
         CIFAR-10 &          LeNet 5 &        8.02\% &        16.49\% &   18.34\% &   82.35\% &       {\bf5.14\%} &         19.39\% &         16.89\% \\
        CIFAR-100 &       ResNet 110 &        35.5\% &         7.03\% &   10.36\% &    10.9\% &         4.74\% &         {\bf2.5\%} &         45.62\% \\
        CIFAR-100 &  ResNet 110 (SD) &       26.42\% &         9.12\% &   10.95\% &    9.12\% &       {\bf8.85\%} &        {\bf8.85\%} &          35.6\% \\
        CIFAR-100 &   Wide ResNet 32 &       33.11\% &         6.22\% &   14.87\% &   11.88\% &       {\bf5.33\%} &          6.31\% &         44.73\% \\
        CIFAR-100 &      DenseNet 40 &       21.52\% &         9.36\% &   10.59\% &  {\bf8.67\%} &         19.4\% &          8.82\% &         38.64\% \\
        CIFAR-100 &          LeNet 5 &       10.25\% &        18.61\% &  {\bf3.64\%} &    9.96\% &         5.22\% &          8.65\% &         18.77\% \\
         ImageNet &     DenseNet 161 &       14.07\% &        13.14\% &   11.57\% &   10.96\% &        12.29\% &        {\bf9.61\%} &              - \\
         ImageNet &       ResNet 152 &        12.2\% &        14.57\% &  {\bf8.74\%} &    8.85\% &        12.29\% &          9.61\% &              - \\
             SVHN &  ResNet 152 (SD) &       19.36\% &        11.16\% &   18.67\% &  {\bf9.09\%} &        18.05\% &         30.78\% &         18.76\% \\
              \midrule
          20 News &            DAN 3 &       17.03\% &        10.47\% &    9.13\% &  {\bf6.28\%} &         8.21\% &          8.24\% &         17.43\% \\
          Reuters &            DAN 3 &     {\bf14.01\%} &        16.78\% &   44.95\% &   36.18\% &        25.46\% &         18.88\% &         19.39\% \\
       SST Binary &         TreeLSTM &       21.66\% &       {\bf3.22\%} &   13.91\% &   36.43\% &         6.03\% &          6.03\% &          6.03\% \\
 SST Fine Grained &         TreeLSTM &       27.85\% &        28.35\% &    19.0\% &  {\bf8.67\%} &        44.75\% &         11.47\% &         11.78\% \\
\bottomrule
\end{tabular}
}

  \caption{MCE (\%) (with $M=15$ bins) on standard vision and NLP datasets before calibration and with various calibration methods. The number following a model's name denotes the network depth. MCE seems very sensitive to the binning scheme and is less suited for small test sets.}
  \label{table:mce}
\end{table*}

\begin{table*}[t!]
	\centering
	\resizebox{\textwidth}{!}{%
\begin{tabular}{c c c c c c c c c}
\toprule
          Dataset &            Model & Uncalibrated & Hist. Binning & Isotonic &      BBQ & Temp. Scaling & Vector Scaling & Matrix Scaling \\
\midrule
            Birds &        ResNet 50 &     {\bf22.54\%} &        55.02\% &   23.37\% &   37.76\% &      {\bf22.54\%} &         22.99\% &         29.51\% \\
             Cars &        ResNet 50 &       14.28\% &        16.24\% &    14.9\% &   19.25\% &        14.28\% &       {\bf14.15\%} &         17.98\% \\
         CIFAR-10 &       ResNet 110 &      {\bf6.21\%} &         6.45\% &    6.36\% &    6.25\% &       {\bf6.21\%} &          6.37\% &          6.42\% \\
         CIFAR-10 &  ResNet 110 (SD) &        5.64\% &         5.59\% &    5.62\% &  {\bf5.55\%} &         5.64\% &          5.62\% &          5.69\% \\
         CIFAR-10 &   Wide ResNet 32 &      {\bf6.96\%} &          7.3\% &    7.01\% &    7.35\% &       {\bf6.96\%} &           7.1\% &          7.27\% \\
         CIFAR-10 &      DenseNet 40 &      {\bf5.91\%} &         6.12\% &    5.96\% &     6.0\% &       {\bf5.91\%} &          5.96\% &           6.0\% \\
         CIFAR-10 &          LeNet 5 &       15.57\% &        15.63\% &   15.69\% &   15.64\% &        15.57\% &       {\bf15.53\%} &         15.81\% \\
        CIFAR-100 &       ResNet 110 &       27.83\% &        34.78\% &   28.41\% &   28.56\% &        27.83\% &       {\bf27.82\%} &         38.77\% \\
        CIFAR-100 &  ResNet 110 (SD) &     {\bf24.91\%} &        33.78\% &   25.42\% &   25.17\% &      {\bf24.91\%} &         24.99\% &         35.09\% \\
        CIFAR-100 &   Wide ResNet 32 &      {\bf28.0\%} &        34.29\% &   28.61\% &   29.08\% &       {\bf28.0\%} &         28.45\% &          37.4\% \\
        CIFAR-100 &      DenseNet 40 &       26.45\% &        34.78\% &   26.73\% &    26.4\% &        26.45\% &       {\bf26.25\%} &         36.14\% \\
        CIFAR-100 &          LeNet 5 &     {\bf44.92\%} &        54.06\% &   45.77\% &   46.82\% &      {\bf44.92\%} &         45.53\% &         52.44\% \\
         ImageNet &     DenseNet 161 &       22.57\% &        48.32\% &    23.2\% &   47.58\% &        22.57\% &       {\bf22.54\%} &              - \\
         ImageNet &       ResNet 152 &     {\bf22.31\%} &         48.1\% &   22.94\% &    47.6\% &      {\bf22.31\%} &         22.56\% &              - \\
             SVHN &  ResNet 152 (SD) &      {\bf1.98\%} &         2.06\% &    2.04\% &    2.04\% &       {\bf1.98\%} &           2.0\% &          2.08\% \\
              \midrule
          20 News &            DAN 3 &       20.06\% &        25.12\% &   20.29\% &   20.81\% &        20.06\% &       {\bf19.89\%} &          22.0\% \\
          Reuters &            DAN 3 &        2.97\% &         7.81\% &    3.52\% &    3.93\% &         2.97\% &        {\bf2.83\%} &          3.52\% \\
       SST Binary &         TreeLSTM &       11.81\% &        12.08\% &   11.75\% &   {\bf11.26\%} &        11.81\% &          11.81\% &       11.81\% \\
 SST Fine Grained &         TreeLSTM &        49.5\% &        49.91\% &   48.55\% &   49.86\% &         49.5\% &         49.77\% &       {\bf48.51\%} \\
\bottomrule
\end{tabular}
}

	\caption{Test error (\%) on standard vision and NLP datasets before calibration and with various calibration methods. The number following a model's name denotes the network depth. Error with temperature scaling is exactly the same as uncalibrated.}
  \label{table:err}
\end{table*}

\begin{table*}[t!]
	\centering
	\resizebox{\textwidth}{!}{%
\begin{tabular}{c c c c c c c c c}
\toprule
          Dataset &            Model & Uncalibrated & Hist. Binning & Isotonic &     BBQ & Temp. Scaling & Vector Scaling & Matrix Scaling \\
\midrule
            Birds &        ResNet 50 &       0.9786 &        1.6226 &   1.4128 &  1.2539 &      {\bf0.8792} &         0.9021 &          2.334 \\
             Cars &        ResNet 50 &       0.5488 &        0.7977 &   0.8793 &  0.6986 &        0.5311 &       {\bf0.5299} &         1.0206 \\
         CIFAR-10 &       ResNet 110 &       0.3285 &        0.2532 &   0.2237 &   0.263 &        0.2102 &         0.2088 &       {\bf0.2048} \\
         CIFAR-10 &  ResNet 110 (SD) &       0.2959 &        0.2027 &   0.1867 &  0.2159 &        0.1718 &       {\bf0.1709} &         0.1766 \\
         CIFAR-10 &   Wide ResNet 32 &       0.3293 &        0.2778 &   0.2428 &  0.2774 &        0.2283 &         0.2275 &       {\bf0.2229} \\
         CIFAR-10 &      DenseNet 40 &       0.2228 &         0.212 &   0.1969 &  0.2087 &      {\bf0.1750} &         0.1757 &          0.176 \\
         CIFAR-10 &          LeNet 5 &       0.4688 &         0.529 &   0.4757 &  0.4984 &         0.459 &       {\bf0.4568} &         0.4607 \\
        CIFAR-100 &       ResNet 110 &       1.4978 &        1.4379 &    1.207 &  1.5466 &      {\bf1.0442} &         1.0485 &         2.5637 \\
        CIFAR-100 &  ResNet 110 (SD) &       1.1157 &        1.1985 &   1.0317 &  1.1982 &      {\bf0.8613} &         0.8655 &         1.8182 \\
        CIFAR-100 &   Wide ResNet 32 &       1.3434 &        1.4499 &   1.2086 &   1.459 &      {\bf1.0565} &         1.0648 &         2.5507 \\
        CIFAR-100 &      DenseNet 40 &       1.0134 &        1.2156 &   1.0615 &  1.1572 &        0.9026 &       {\bf0.9011} &         1.9639 \\
        CIFAR-100 &          LeNet 5 &       1.6639 &        2.2574 &   1.8173 &  1.9893 &      {\bf1.6560} &         1.6648 &         2.1405 \\
         ImageNet &     DenseNet 161 &       0.9338 &        1.4716 &   1.1912 &  1.4272 &        0.8885 &       {\bf0.8879} &              - \\
         ImageNet &       ResNet 152 &       0.8961 &        1.4507 &   1.1859 &  1.3987 &      {\bf0.8657} &         0.8742 &              - \\
             SVHN &  ResNet 152 (SD) &       0.0842 &        0.1137 &    0.095 &  0.1062 &      {\bf0.0821} &         0.0844 &         0.0924 \\
              \midrule
          20 News &            DAN 3 &       0.7949 &        1.0499 &   0.8968 &  0.9519 &        0.7387 &       {\bf0.7296} &         0.9089 \\
          Reuters &            DAN 3 &        0.102 &        0.2403 &   0.1475 &  0.1167 &        0.0994 &       {\bf0.0990} &         0.1491 \\
       SST Binary &         TreeLSTM &       0.3367 &        0.2842 &   0.2908 &  0.2778 &        {\bf0.2739} &       {\bf0.2739} &        {\bf0.2739} \\
 SST Fine Grained &         TreeLSTM &       1.1475 &        1.1717 &   1.1661 &   1.149 &        1.1168 &       {\bf1.1085} &         1.1112 \\
\bottomrule
\end{tabular}
}

	\caption{NLL (\%) on standard vision and NLP datasets before calibration and with various calibration methods. The number following a model's name denotes the network depth. To summarize, NLL roughly follows the trends of ECE.}
  \label{table:nll}
\end{table*}

\section{Additional Reliability Diagrams}
\label{sup:reliability}

\begin{figure*}[h!]
  \centering
  \includegraphics[width=\textwidth]{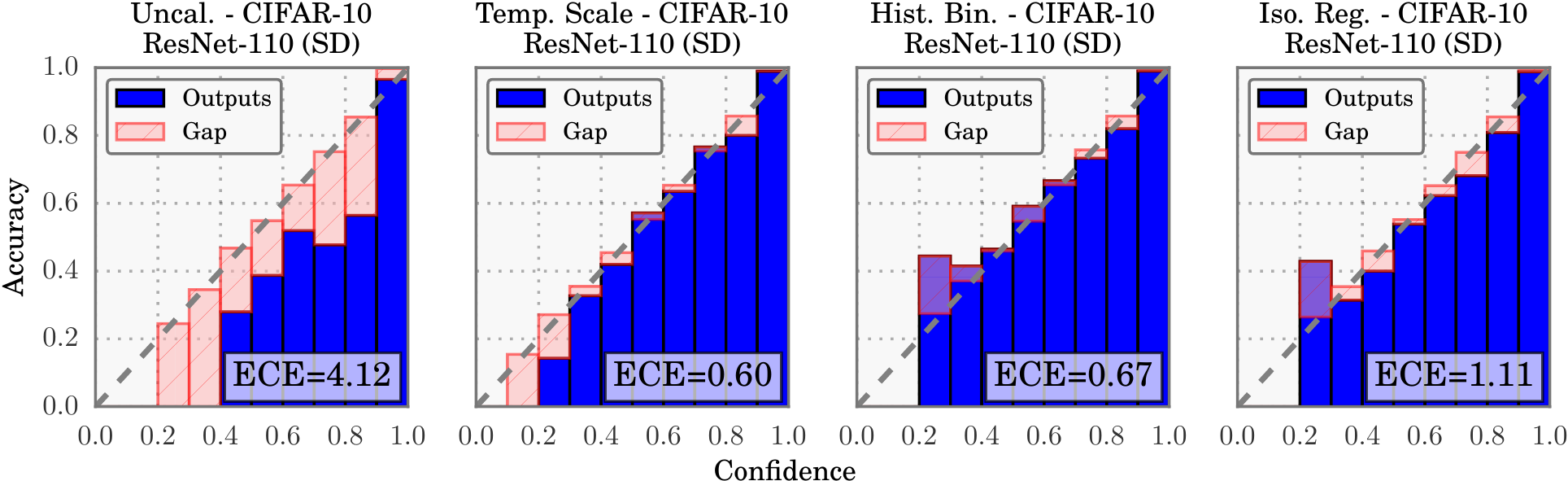}
  \caption{Reliability diagrams for CIFAR-10 before (far left) and after calibration (middle left, middle right, far right).}
  \label{figure.reliability-cifar10}
\end{figure*}
\begin{figure*}[h!]
  \centering
  \includegraphics[width=\textwidth]{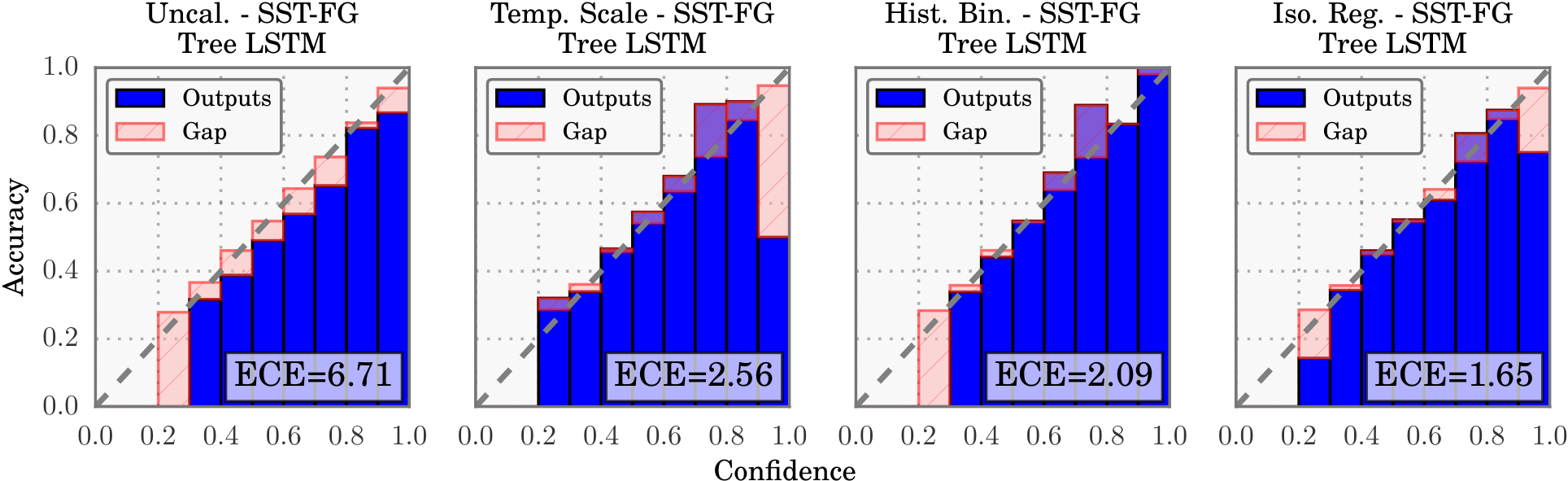}
  \caption{Reliability diagrams for SST Binary and SST Fine Grained before (far left) and after calibration (middle left, middle right, far right).}
  \label{figure.reliability-sst_fg}
\end{figure*}
\begin{figure*}[h!]
  \centering
  \includegraphics[width=\textwidth]{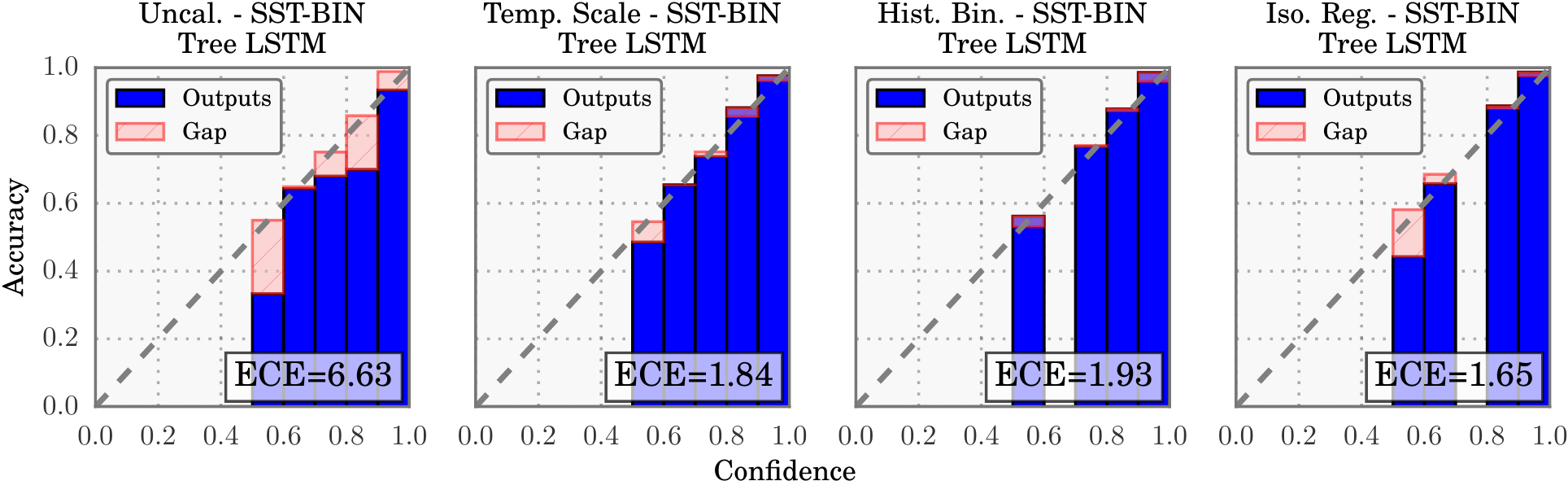}
  \caption{Reliability diagrams for SST Binary and SST Fine Grained before (far left) and after calibration (middle left, middle right, far right).}
  \label{figure.reliability-sst_binary}
\end{figure*}

We include reliability diagrams for additional datasets: CIFAR-10 (\autoref{figure.reliability-cifar10}) and SST (\autoref{figure.reliability-sst_fg} and \autoref{figure.reliability-sst_binary}). Note that, as mentioned in \autoref{sec:definitions}, the reliability diagrams do not represent the proportion of predictions that belong to a given bin.

\end{document}